%% file: paper.tex
\documentclass[letterpaper]{article}
\usepackage{uai2020}
\usepackage[margin=1in]{geometry}

\usepackage{times}
\usepackage{graphicx}
\usepackage{natbib}
\author{} 

\author{ {\bf Bhairav Mehta\thanks{~~Correspondence to \texttt{mehtabha@mila.quebec}}} \\
Université de Montréal, Mila
\And
{\bf Tristan Deleu}\thanks{~~Denotes equal contribution} \\
Université de Montréal, Mila \\
\And
{\bf Sharath Chandra Raparthy\footnotemark[2]} \\
Mila \\ 
\AND 
{\bf Chris J. Pal}   \\
Polytechnique Montréal, Mila \\
CIFAR AI Chair
\And 
{\bf Liam Paull}   \\
Université de Montréal, Mila \\
CIFAR AI Chair
}

\usepackage{booktabs}
\usepackage{graphicx}
\usepackage{algorithm}
\usepackage{algorithmic}
\usepackage{amsmath}

\DeclareMathOperator*{\argmin}{arg\,min}

\usepackage{xcolor}

\title{Curriculum in Gradient-Based Meta-Reinforcement Learning}
\begin{document}

\maketitle


\begin{abstract}  
Gradient-based meta-learners such as Model-Agnostic Meta-Learning (MAML) have shown strong few-shot performance in supervised and reinforcement learning settings. However, specifically in the case of meta-reinforcement learning (meta-RL), we can show that gradient-based meta-learners are sensitive to task distributions.  With the wrong curriculum, agents suffer the effects of \textit{meta-overfitting}, shallow adaptation, and adaptation instability. In this work, we begin by highlighting intriguing failure cases of gradient-based meta-RL and show that task distributions can wildly affect algorithmic outputs, stability, and performance.
To address this problem, we leverage insights from recent literature on \textit{domain randomization} and propose meta Active Domain Randomization (meta-ADR), which learns a curriculum of tasks for gradient-based meta-RL in a similar as ADR does for sim2real transfer. We show that this approach induces more stable policies on a variety of simulated locomotion and navigation tasks. We assess in- and out-of-distribution generalization and find that the learned task distributions, even in an unstructured task space, greatly improve the adaptation performance of MAML. 
Finally, we  motivate the need for better benchmarking in meta-RL that prioritizes \textit{generalization} over single-task adaption performance.
\end{abstract}

\input{samplebody-conf}

\bibliography{paper}  

\end{document}

%% file: samplebody-conf.tex
\begin{figure}[tb]
   \includegraphics[width=3.2in]{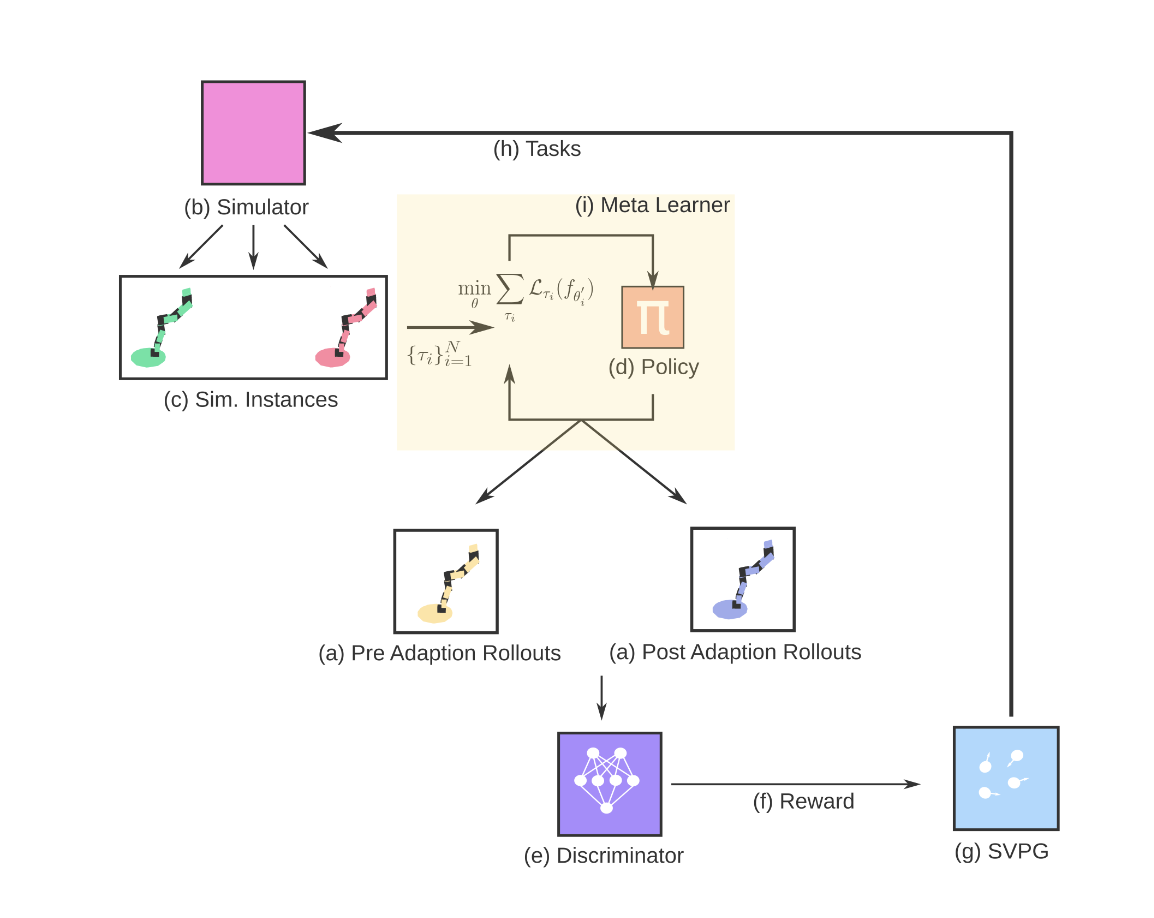}
   \caption{Meta-ADR proposes tasks to a meta-RL agent, helping learn a curriculum of tasks rather than uniformly sampling them from a set distribution. A \textbf{discriminator} learns a \textbf{reward} as a proxy for task-difficulty, using pre- and post-adaptation rollouts as input. The reward is used to train \textbf{SVPG particles}, which find the tasks causing the meta-learner the most difficulty after adaption. The particles propose a diverse set of tasks, trying to find the tasks that are currently causing the agent the most difficulty.
    }
    \label{fig:overview}
\end{figure}

\section{Introduction}

Meta-learning concerns building models or agents that can learn how to adapt quickly to new tasks from datasets which are orders of magnitudes smaller than their standard supervised learning counterparts. Put differently, meta-learning concerns learning \textit{how to learn}, rather than simply maximizing performance on a single task or dataset. Gradient-based meta-learning has seen a surge of interest, with the foremost algorithm being Model-Agnostic Meta-Learning (MAML) \citep{maml17}. Gradient-based meta-learners are fully trainable via gradient descent, and have shown strong performance on various supervised and reinforcement learning tasks \citep{maml17, rakelly2019efficient}.

The focus of this work is on an understudied hyperparameter within the gradient-based meta-learning framework: the distribution of tasks. In MAML, this distribution is assumed given and is used to sample tasks for meta-training of the MAML agent. In supervised learning, this quantity is relatively well-defined, as we often have a large dataset for a task such as image classification \footnotemark \footnotetext{Even in regression, a function such as a sinusoid is often provided by the experimenter as the task distribution.}. As a result, the distribution of tasks, $p(\tau)$, is built from random minibatches sampled from this distribution.

However, in the meta-reinforcement learning setting, this task distribution is poorly defined, and is often handcrafted by a human experimenter with the target task in mind. While the task samples themselves are pulled randomly from a range or distribution (i.e a locomoter asked to achieve a target velocity), the distribution \textit{itself} needs to be specified. 
In practice, the distribution $p(\tau)$ turns out to be an extremely sensitive hyperparameter in Meta-RL: too "wide" of a distribution (i.e the variety of tasks is too large) leads to underfitting, with agents unable to specialize to the given target task even with larger numbers of gradient steps; too "narrow", and we see poor generalization and adaption to even slightly out-of-distribution environments. 


Even worse, randomly sampling (as is often the case) from $p(\tau)$ can allow for sampling of tasks that can cause interference and optimization difficulties, especially when tasks are qualitatively different (due to difficulty or task definitions being changed too much by the physical parameters that are varied).

This phenomena, called \textit{meta-overfitting} (or meta-underfitting, in the former, "wide" case), is not new to recent deep reinforcement learning problem settings. Domain randomization \citep{tobin2017domain}, a popular \textit{sim2real} transfer method, faces many of the same issues when learning robotic policies purely in simulation. Here, we will show that meta-reinforcement learning has the analogous issues regarding generalization, which we can attribute to the random sampling of tasks. We will then describe the repurposing of a recent algorithm called \textit{Active Domain Randomization} \citep{mehtaadr19}, which aims to learn a curriculum of tasks in unstructured task spaces. In this work, we address the problem of meta-overfitting by explicitly optimizing for the task distribution represented by $p(\tau)$. The incorporation of a learned curriculum leads to stronger generalization performance and more robust optimization. Our results highlight the need for continued work in analysis of the effect of task distributions on meta-RL performance and underscoring the potential for curriculum learning techniques.



\section{Background}

In this section we briefly cover reinforcement learning, meta-learning, and curriculum learning ideas touched upon in later parts of the paper. 

\subsection{Reinforcement Learning}
We consider a reinforcement learning setting where a task $\tau$ is defined as a Markov Decision Process (MDP), a tuple $(S, A, T, R, \gamma)$ where  $S$ is the state space, $A$ is the action space, $T$ is transition function $ T: S\: \times A \rightarrow S$, $R$ is a reward function and $\gamma$ is a \textit{discount factor} which lives within $(0, 1)$. The goal of reinforcement learning is to learn a function $\pi$ parameterized by $\theta$ in such a way that it maximizes the expected total discounted reward \citep{sutton2018reinforcement}.

\subsection{Meta-Learning}
Most deep learning models are built to solve only one task, and often lack the ability to generalize and quickly adapt to solve a new set of tasks. Meta-learning involves learning a \textit{learning algorithm} which can adapt quickly rather than learning from scratch. Several methods have been proposed, treating the learning algorithm as a recurrent model capable of remembering past experience \citep{santoro2016meta,munkhdalai2017metanetworks,mishra2017snail}, as a non-parametric model \citep{koch2015siamese,matchingnet2016,snell2017protonet}, or as an optimization problem \citep{ravi2016optimization,maml17}. In this paper, we focus on a popular version of a gradient-based meta-learning algorithm called Model Agnostic Meta Learning (MAML; \citep{maml17}).

\subsubsection{Gradient-based Meta-Learning}

The main idea in MAML is to find a good parameter initialization such that the model can adapt to a new task, $\tau$, quickly. Formally, given a distribution of tasks $p(\tau)$ and a loss function ${\mathcal{L}_{\tau}}$
 corresponding to each task, the aim is to find parameters $\theta$ such that the model $f_\theta$ can adapt to new tasks with one or few gradient steps. For example, in the case of a single gradient step, the parameters $\theta'_{\tau}$ adapted to the task $\tau$ are
 \begin{equation}
     \theta'_{\tau} = \theta - \alpha \nabla_{\theta}\mathcal{L}_{\tau}(\mathcal{D}_{\mathrm{train}}, f_{\theta}),
     \label{eq:innerloop}
 \end{equation}
 with step size $\alpha$, where the loss is evaluated on a (typically small) dataset $\mathcal{D}_{\mathrm{train}}$ of training examples from task $\tau$. In order to find a good initial value of the parameters $\theta$, the objective function being optimized in MAML is written as 
\begin{equation}
    \min_\theta \sum_{\tau_i} \mathcal{L}_{\tau_i}(\mathcal{D}_{\mathrm{test}}, f_{\theta'_{\tau_{i}}}),
\end{equation}
where it evaluates the performance in generalization on some test examples $\mathcal{D}_{\mathrm{test}}$ for task $\tau$. The meta objective function is optimized by gradient descent where the parameters are updated according to
\begin{equation}
    \theta \leftarrow \theta - \beta \nabla_{\theta}\sum_{\tau_i}\mathcal{L}_{\tau_i}(\mathcal{D}_{\mathrm{test}}, f_{\theta'_{\tau_{i}}}),
    \label{eq:metaobjective}
\end{equation}
where $\beta$ is the outer step size.

\subsection{Meta-Reinforcement Learning}
In addition to few-shot supervised learning problems, where the number of training examples is small, meta-learning has also been successfully applied to reinforcement learning problems. In meta-reinforcement learning, the goal is to find a policy that can quickly adapt to new environments, generally from only a few trajectories. \cite{rakelly2019efficient} treat this problem by conditioning the policy on a latent representation of the task, and \cite{duan2016rl2,wang2016learning} represent the reinforcement learning algorithm as a recurrent network, inspired by the ``black-box'' meta-learning methods mentioned above. Some meta-learning algorithms can even be adapted to reinforcement learning with minimal changes \citep{mishra2017snail}. In particular, MAML has also shown some success on robotics applications \citep{finn2017oneshot}. In the context of reinforcement learning, $\mathcal{D}_{\mathrm{train}}$ and $\mathcal{D}_{\mathrm{test}}$ are datasets of trajectories sampled by the policies before and after adaptation (i.e rollouts in $\mathcal{D}_{\mathrm{train}}$ are sampled before the gradient step in Equation \ref{eq:innerloop}, whereas those in $\mathcal{D}_{\mathrm{test}}$ are sampled after). The loss function used for the adaptation is REINFORCE \citep{williams92reinforce}, and the outer, meta objective in Equation \ref{eq:metaobjective} is optimized using TRPO \citep{schulman2015trust}.


\subsection{Active Domain Randomization}

Active Domain Randomization (ADR) \citep{mehtaadr19} builds upon the framework of Domain Randomization \citep{tobin2017domain}. Domain randomization, a useful, zero-shot technique for transferring robot policies from simulation to real hardware, uniformly samples randomized environments (effectively, tasks) that an agent must solve. ADR improves DR by learning an active policy that proposes a curriculum of tasks to train an inner-loop, black-box agent. ADR, mostly used in the zero-shot learning scenario of \textit{simulation-to-real transfer}, uses Stein Variational Policy Gradient (SVPG) \citep{svpg} to learn a set of parameterized particles, $\{ \mu_{\phi_i} \}_{i=1}^N$ that proposes randomized environments which are subsequently used to train the agent. 



SVPG benefits from both the Maximum-Entropy RL framework \citep{ziebart2010modeling} and a kernel term that repulses similar policies to encourage particle, and therefore task, diversity. This allows SVPG to hone in on regions of high reward while maintaining variety, which allows ADR to outperform many existing methods in terms of performance and generalization on zero-shot learning and robotic benchmarks. 

To train the particles, ADR uses a discriminator to distinguish between trajectories generated in a proposed randomized environment and those generated by the same policy in a default, reference environment. Intuitively, ADR is optimized to find environments where the same policy produces different behavior in the two types of environments, signalling a probable weakness in the policy when evaluated on those types of randomized environments.


\section{Related Work}
\label{sec:related-work}
When discussing meta-reinforcement learning, to the best of our knowledge, the task distribution $p(\tau)$ has never been studied or ablated upon. As most benchmark environments and tasks in meta-RL stem from two papers (\citep{maml17, rothfuss2018promp}, with the task distributions being prescribed with the environments), the discussion in meta-RL papers has almost always centered around the computation of the updates \citep{rothfuss2018promp}, practical improvements and approximations made to improve efficiency, or learning exploration policies with meta-learning \citep{stadie2018considerations, gurumurthy2019mame, gupta2018metareinforcement}. In this section, we briefly discuss prior work in curriculum learning that bears the most similarity to the analyses we conduct here.

Starting with the seminal curriculum learning paper \citep{curriculumbengio09}, many different proposals to learn an optimal \textit{ordering} over tasks has been studied. Curriculum learning has been tackled with Bayesian Optimization \citep{bayesoptCL}, multi-armed bandits \citep{graves2017automated}, and evolutionary strategies \citep{wang2019paired} in supervised learning and reinforcement learning settings, but here, we focus on the latter. However, in most work, the task space is almost always discrete, with a teacher agent looking to choose the best next task over a set of $N$ pre-made tasks. The notion of \textit{best} has also been explored in depth, with metrics being based on a variety of things from ground-truth accuracy or reward to adversarial gains between a teacher and student agent \citep{pinto2017robust}.

However, up until recently, the notion of continuously-parameterized curriculum learning has been studied less often. Often, continuous-task curriculum learning exploits a notion of \textit{difficulty} in the task itself. In order to get agents to hop over large gaps, it's been empirically easier to get them to jump over smaller ones first \citep{heess2017emergence}; likewise, in navigation domains, its been easier to show easier goals and \textit{grow} a goal space \citep{pong2019skewfit}, or even work backwards towards the start state in a reverse curriculum manner \citep{florensa2017reverse}.

While deep reinforcement learning, particularly in robotics, has a seen a large amount of curriculum learning papers in recent times \citep{mehtaadr19, OpenAI2019SolvingRC}, curriculum learning has not been extensively researched in meta-RL. This may be partly due to the naissance of the field; only recently was a large-scale, multi-task benchmark for meta-RL released \citep{yu2019metaworld}. As we hope to show in this work, the notions of tasks, task distributions, and curricula in meta-learning are fruitful avenues of study, and can make (or break) many of the meta-learning algorithms in use today.

\section{Motivation}
\label{sec:motivation}

We begin with a simple question: 

\begin{center}
    \textit{Does the meta-training task distribution in meta-RL really matter?}
\end{center}

To answer this question, we run a standard meta-reinforcement learning benchmark, \textit{2D-Navigation-Dense}. In this environment, a point-mass must navigate to a goal, with rewards given at each timestep proportional to the Euclidean distance between the goal and the current position. 

We take the hyperparameters and experiment setup from the original MAML work and simply change \textit{the task distribution} from which the 2D goal is \textit{uniformly} sampled. We then show generalization results of the final, meta-learned initial policy \textit{after a single gradient step}. We then track the generalization of the one-step adaptation performance across a wide range of target goals.

\begin{figure}[tb]
    \centering
    \includegraphics[width=3.2in]{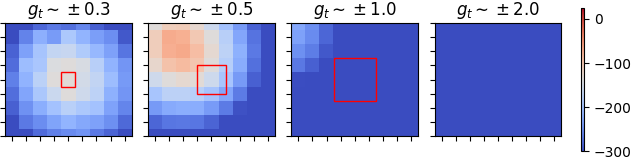}
    \caption{Various agents' final adaption to a range of target tasks. The agents vary only in training task distributions, shown as red overlaid boxes. \textbf{Redder is higher reward.}}
    \label{fig:motivation2d}
\end{figure}

In \textit{2D-Navigation-Dense}, the training distribution prescribes goals where each coordinate is traditionally sampled between $[-0.5, 0.5]$ (the second plot in Figure \ref{fig:motivation2d})  with the agent always beginning at $[0, 0]$. We then evaluate each goal in the grid between $[-2, 2]$ at 0.5 intervals, allowing us to test both in- and out-of-distribution generalization.

We see from Figure \ref{fig:motivation2d} an interesting phenomenon, particularly as the training environment shifts away from the one which samples goal coordinates $g_t \sim [-0.5, 0.5]$. While the standard environment from \citep{maml17} generalizes reasonably well, shifting the training distribution even slightly ruins generalization of the adapted policy. What's more, when shown the entire test distribution, MAML fails to generalize to it. We see that on even the simplest environments, the meta-training task distribution seems to have a profound effect, motivating the need for dedicating more attention towards selecting the task distribution $p(\tau)$.

Upon further inspection, we find that shifting the meta-training distribution destabilizes MAML, leading to poor performance when averaged. The first environment, where $g_t \sim [-0.3, 0.3]$ has three out of five random seeds that converge, with the latter two, $g_t \sim [-1.0, 1.0]$ and $g_t \sim [-2.0, 2.0]$, have two and one seeds that converge respectively. The original task distribution sees convergence in all five random seeds tested, hinting at a difference in stability due to the goals, and therefore task distribution, that each agent sees. This hints at a hidden importance of the task distribution $p(\tau)$, a hypothesis we explore in greater detail in the next section.


\section{Method}
\label{sec:method}
As we saw in the previous section, uniformly sampling tasks from a set task distributions highly affects generalization performance of the resulting meta-learning agent. Consequently, in this work, we optimize for a curriculum over the task distribution $p(\tau)$:


\begin{equation}
    \argmin_{\tau_i \sim p(\tau)} \min_{\theta} \sum_{\tau_i} \mathcal{L}_{\tau_i} (f_{\theta_{i}^{'}} )
\end{equation}

where $\theta_{i}^{'}$ are the updated parameters after a single meta-gradient update.

While curriculum learning has had success in scenarios where task-spaces are structured, learning curricula in unstructured task-spaces, where an intuitive scale of \textit{difficulty} might be lacking, is an understudied topic. 
However, learning such curricula has seen a surge of interest in the problem of \textit{simulation transfer} in robotics, where policies trained in simulation are transferred zero-shot (no fine-tuning) for use on real hardware. Using a method called domain randomization ~\citep{tobin2017domain}, several recent methods ~\citep{OpenAI2019SolvingRC,mozifandomainrand19} propose how to learn a curriculum of \textit{randomizations} - which randomized environments would be most useful to show the learning agent in order to make progress on the held-out target task: the real robot. 

In the meta-RL setting, the learned curriculum would be over the space of tasks. For example,  in \textit{2D-Navigation-Dense}, this would be where goals are sampled, or in \textit{HalfCheetahVelocity}, another popular meta-RL benchmark, the goal velocity the locomotor must achieve. 

As learning the curriculum is often treated as a reinforcement learning problem, it requires a reward in order to calculate policy gradients. While many of the methods from the domain randomization literature use proxies such as completion rates or average reward, the optimization scheme depends on the reward function of the \textit{task}. In meta-learning, optimization and reward maximization on a \textit{single} task is not the goal, and such an approach may lead to counter-intuitive results.

A more natural fit in the meta-learning scenario would be to somehow use the \textit{qualitative difference} between the pre- and post-adaptation trajectories. Like a good teacher with a struggling student, the curriculum could shift towards where the meta-learner needs help. For example, tasks in which \textit{negative adaptation} \citep{deleunegativeadapt19} occurs, or where the return from a pre-adapted agent is higher the post-adapted agent, would be prime tasks to focus on for training.

To this end, we modify \textit{Active Domain Randomization} (ADR) to calculate such a score between the two types of trajectories. Rather than using a reference environment as in ADR, we ask a discriminator to differentiate between the pre- and post-adaptation trajectories. If a particular task generates trajectories that can be distinguished by the discriminator after adaptation, we focus more heavily on these tasks by providing the high-level optimizer, parameterized by Stein Variational Policy Gradient, a higher reward. 

Concretely, we provide the particles the reward:

\begin{equation}
  r_i = \log(f_\psi(y|D_i)) 
  \label{eq:discrimreward}
\end{equation}

\noindent
where discriminator $f_\psi$ produces a boolean prediction of whether the trajectory $D_i$ is a pre-adaptation ($y = 0$) or post-adaptation ($y = 1$) trajectory. 
We present the algorithm, which we term Meta-ADR, in Algorithm \ref{alg:adr}.
\begin{algorithm}[tb]
   \caption{Meta-ADR}
   \label{alg:adr}
\begin{algorithmic}[1]
   \STATE {\textbf{Input} Task distribution $p(\tau)$} 
   \STATE \textbf{Initialize} $\pi_\theta$: agent policy, $\mu_{\phi}$: SVPG particles, $f_{\psi}$: discriminator
   \WHILE{ \textbf{not} $max\_epochs$}
   \FOR{ \textbf{each} particle $\mu_\phi$}
   \STATE \textbf{sample tasks} $\tau_i \sim \mu_\phi(\cdot)$, bounded by support of $p(\tau)$
   \ENDFOR
   \FOR{\textbf{each} $\tau_i$}
   \STATE $D_{pre}$, $D_{post}$ = MAML$_{RL}$($\pi_\theta, \tau_i$)
   \STATE{Calculate $r_i$ for $\tau_i$ using $D_{post}$ (Eq. (\ref{eq:discrimreward}))}
   \ENDFOR
   \STATE \textit{// Gradient Updates}
   \STATE Update particles using SVPG update rule and $r_i$
   \STATE Update $f_\psi$ with $D_{pre}$ and $D_{post}$ using SGD.
   \ENDWHILE
\end{algorithmic}
\end{algorithm}
Meta-ADR learns a curriculum in this unstructured task space without relying on the notion of task performance or reward functions. Note that Meta-ADR runs the original MAML algorithm as a subroutine, but in fact
Meta-ADR can run any meta-learning subroutine (i.e Reptile \citep{nichol2018firstorder}, PEARL \citep{rakelly2019efficient}, or First-Order MAML). In this work, we abstract away the meta-learning subroutine, focusing instead on the effect of task distributions on the learner's generalization capabilities. An advantage of Meta-ADR over the ADR original formulation is that unlike ADR, Meta-ADR requires no additional rollouts, using the rollouts already required by gradient-based meta-reinforcement learners to optimize the curriculum. 

\section{Results}
\label{sec:results}

In this section, we show the results of uniform sampling of the standard MAML agent when changing task distribution $p(\tau)$, while also benchmarking against a MAML agent trained with a learned task distribution using Meta-ADR. All hyperparameters for each task are taken from \citep{maml17, rothfuss2018promp}, with the exception that we take the \textit{final} policy at the end of 200 meta-training epochs instead of the best-performing policy over 500 meta-training epochs. We use the code from \citep{deleu2019mamlrl} to run all of our experiments. Unless otherwise noted, all experiments are run and averaged across five random seeds. All results are shown after a single gradient step during meta-test time. For each task, we artificially create a generalization range; potentially disjoint from the training distribution of target goals, velocities, headings, etc., and we evaluate each agent both in- and out-of-distribution.

Importantly, since our focus is on generalization, we \textbf{evaluate the final policy}, rather than the standard, \textit{best-performing} policy. As MAML produces a final \textit{initial} policy, when evaluating for generalization for meta-learning, we adapt that initial policy to each target task, and report the adaption results. In addition, in certain sections, we discuss \textit{negative} adaption, which is simply the performance difference between the final, adapted policy and the final, initial policy. When this quantity is negative, as noted in \citep{deleunegativeadapt19}, we say that the policy has \textit{negatively} adapted to the task.

We present results from standard Meta-RL benchmarks in Sections \ref{sec:navigation} and \ref{sec:locomotion}, and in general find that Meta-ADR stabilizes the adaption procedure. However, this finding is not universal, as we note in Section \ref{sec:failurecases}. 

In Subsections \ref{sec:failurecases}, we highlight a need for better benchmarking and failure cases (overfitting and biased, non-uniform generalization) that both Meta-ADR and uniform-sampling methods seem to suffer from.

\subsection{Navigation}
\label{sec:navigation}

In this section we evaluate meta-ADR on two navigation tasks: 2D-Navigation-Dense and Ant-Navigation. 

\subsubsection{2D-Navigation-Dense}
\label{sec:2dnav-dense-results}

We train the \textit{same} meta-learning agent from Section \ref{sec:motivation} on \textit{2D-Navigation-Dense}, except this time we use the tasks\footnotemark\footnotetext{In navigation environments, tasks are parameterized by the \texttt{x, y} location of the goal.} proposed by Meta-ADR, using a learned curriculum to propose the next best task for the agent. We evaluate generalization across the same scaled up square spanning the ranges of $[-2, 2]$ in both dimensions. 

\begin{figure}[tb]
    \centering
    \includegraphics[width=1.0\columnwidth]{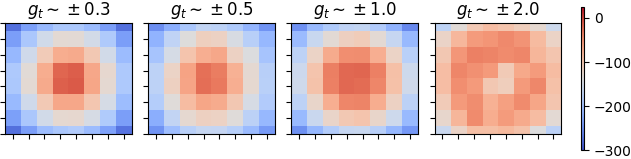}
    \caption{When a curriculum of tasks is learned with Meta-ADR, we see the stability of MAML improve. \textbf{Redder is higher reward.}}
    \label{fig:adr2d}
\end{figure}

From Figure \ref{fig:adr2d}, we see that, with a learned curriculum, the agent generalizes much better, especially \textit{within} the training distribution. A MAML agent trained with a Meta-ADR curriculum also generalizes out-of-distribution with much stronger performance. These results hint at the strong dependence of MAML performance and the task distribution $p(\tau)$, especially when compared to those in Figure \ref{fig:motivation2d}. Learning such a task curriculum with a method such as Meta-ADR helps alleviate some instability.

\vspace{-7pt}
\subsubsection{Ant-Navigation}
\label{sec:ant-navigation-results}

Interestingly, on a more complex variant of the same task, \textit{Ant-Navigation}, the benefits of such a learned curriculum are minimized. 
In this task, an eight-legged locomoter is tasked with achieving goal positions sampled from a predetermined range; the standard environment samples the goal positions from a box centered at $(0, 0)$, with each coordinate sampled from $g \sim [-3, 3]$. 
We systematically evaluate each agent on a grid with both axes ranging between $[-7, 7]$, with a $0.5$ step interval.

\begin{figure}[ht]
    \centering
    \includegraphics[width=1.0\columnwidth]{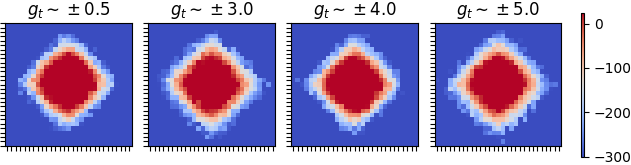}
    \includegraphics[width=1.0\columnwidth]{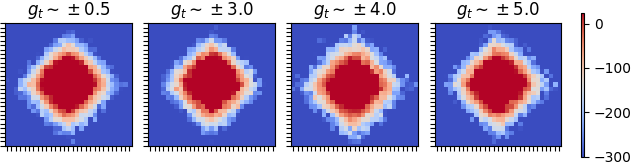}
    \caption{In the Ant-Navigation task, both uniformly sampled goals (top) and a learned curriculum of goals with Meta-ADR (bottom) are stable in performance. We attribute this to the extra components in the reward function. Redder is higher reward.}
    \label{fig:antnav}
\end{figure}

In Figure \ref{fig:antnav}, we qualitatively see the same generalization across all training task distributions when comparing a randomly sampled task curriculum and a learned one. We hypothesize that this stability comes mainly from the control components of the reward, leading to a smoother, stabler performance across all training distributions. In addition, generalization is unaffected by the choice of distribution, pointing to differences between this task and the simpler version.
 
Compared to the \textit{2D-Navigation-Dense}, \textit{Ant-Navigation} also receives a dense reward related to its distance to target, specifically ``control cost, a contact cost, a survival reward, and a penalty equal to its L1 distance to the target position.'' In comparison, the \textit{2D-Navigation-Dense} task, while a simpler control problem, receives reward information only related to the Euclidean distance to the goal. Counter-intuitively, this simplicity results in \textit{less} stable performance when uniformly sampling tasks, an ablation which we hope to study in future work.

\subsection{Locomotion}
\label{sec:locomotion}
We now consider \textit{locomotion}, another standard meta-RL benchmark, where we are tasked with training an agent to quickly move in a particular target velocity (Section \ref{sec:targetvel}) or in a particular direction (Section \ref{sec:humanoiddirectional-results}). In this section, we focus on two high-dimensional continuous control problems. In the \textit{AntVelocity}, an eight-legged locomoter must run at a specific speed, with the task space (both for learned and random curricula) being the target velocity. In \textit{Humanoid-Direc-2D}, a benchmark introduced by \citep{rothfuss2018promp}, an agent must learn to run in a target direction, $\theta$ in a 2D plane. 

Both tasks are extremely high-dimensional in both observation and action space. The ant has a $(111 \times 1)$ sized observation space, with each step requiring an action vector of length eight. The Humanoid, which takes in a $(376 \times 1)$ element state, requires an action vector of length 17.

\subsubsection{Target Velocity Tasks}
\label{sec:targetvel}

\begin{figure}[h]
    \centering
    \includegraphics[width=0.45\columnwidth]{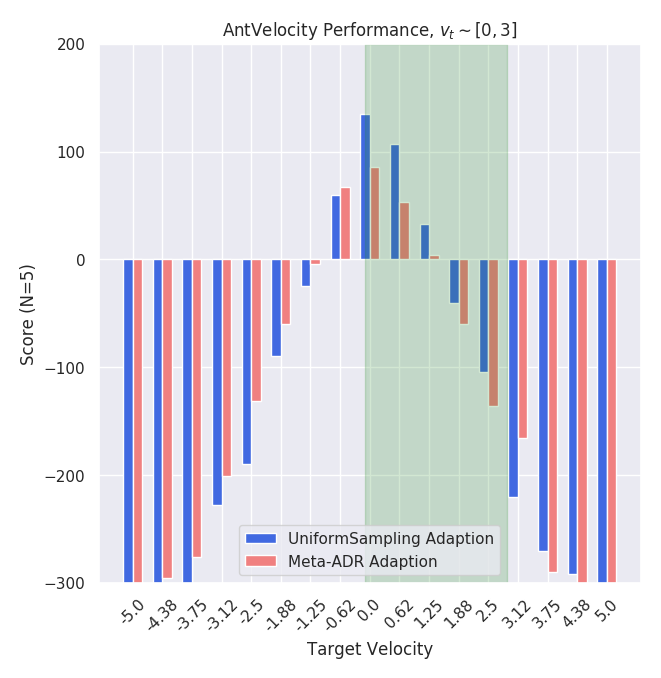}
    \includegraphics[width=0.45\columnwidth]{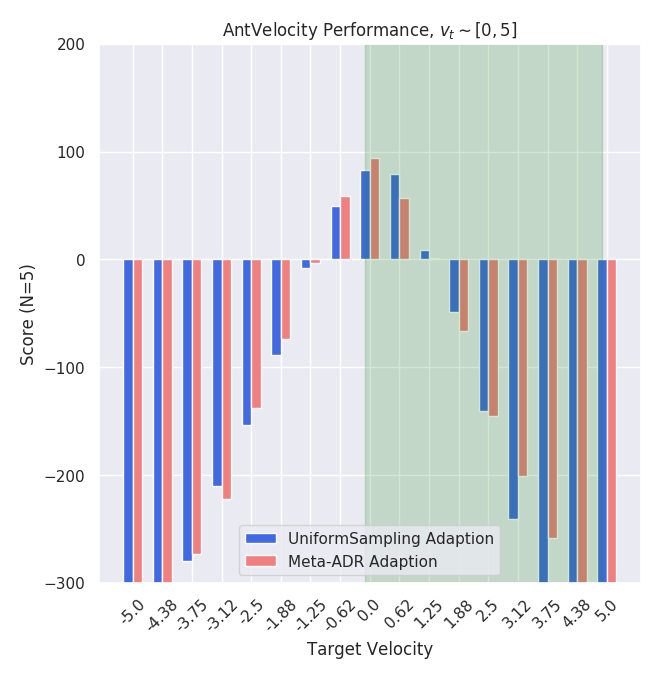}
    \caption{Ant-Velocity sees less of a benefit from curriculum, but performance is greatly affected by a correctly-calibrated task distribution (left). In a miscalibrated one (right), we see that performance from a learned curriculum is slightly more stable.}
    \label{fig:antvel}
\end{figure}

When dealing with the target velocity task, we train an ant locomoter to attain target speeds sampled from $v_t \sim [0, 3]$ (Figure \ref{fig:antvel} left, the standard variant of \textit{AntVelocity}) and speeds sampled from $v_t \sim [0, 5]$ (Figure \ref{fig:antvel} right). 

While we see that learned curricula make insignificant amounts of performance \textit{improvement} over random sampling when shown the same task distribution, we see large differences in performance between task distributions, motivating our hypothesis that $p(\tau)$ is a crucial hyperparameter for successful meta-RL. In additon, we notice that the highest scores are attained on the velocities closer to the easiest variant of the task: a $v_t = 0$, which requires the locomoter to stand completely still. We expand on this oddity in Section \ref{sec:randomsampling}. 

\subsubsection{Humanoid Directional}
\label{sec:humanoiddirectional-results}


\begin{figure}[h]
    \centering
    \includegraphics[width=1.0\columnwidth]{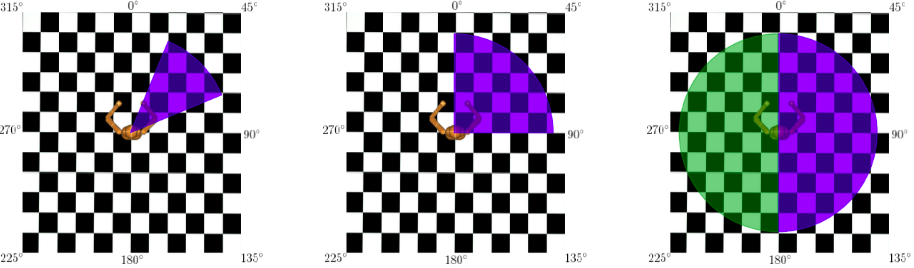}
    \caption{In the high-dimensional Humanoid Directional task, we evaluate many different training distributions to understand the effect of $p(\tau)$ on generalization in difficult continuous control environments. In particular, we focus on \textit{symmetric} variants of tasks - task distributions that mirror each other, such as $0-\pi$ and $\pi-2\pi$ in the right panel. Intuitively, when averaged over many trials, such mirrored distrbutions should produce similar trends of in and out-of-distribution generalization.}
    \label{fig:humanoid-directional-description}
\end{figure}

In the standard variant of \textit{Humanoid-Direc-2D}, a locomoter is tasked with running in a particular direction, sampled from $[0, 2\pi]$. This task makes no distinction regarding target velocity, but rather calculates the reward based on the agent's heading and other control costs. 

In this task, we shift the distribution from $[0, 2\pi]$ to subsets of this range, subsequently training and evaluating MAML agents across the entire range of tasks between $[0, 2\pi]$,  as seen in the first two panels of Figure \ref{fig:humanoid-directional-description}. Again, we compare agents trained with the standard uniformly-random sampled task distribution against those trained with a learned curriculum using Meta-ADR.

\begin{figure}[h]
    \centering
    \includegraphics[width=0.45\columnwidth]{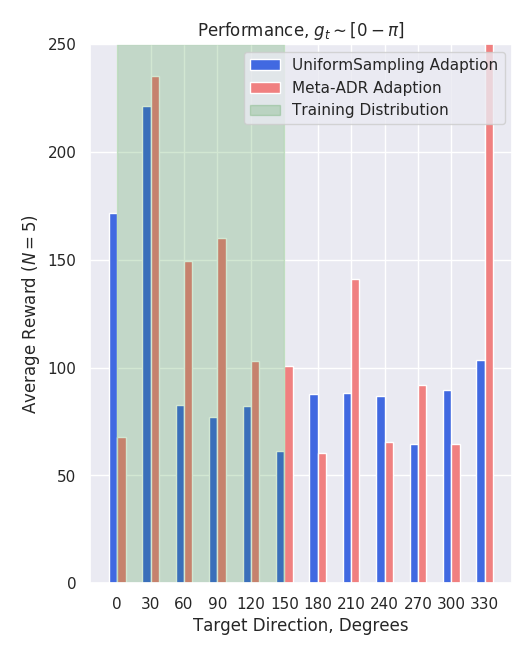}
    \includegraphics[width=0.45\columnwidth]{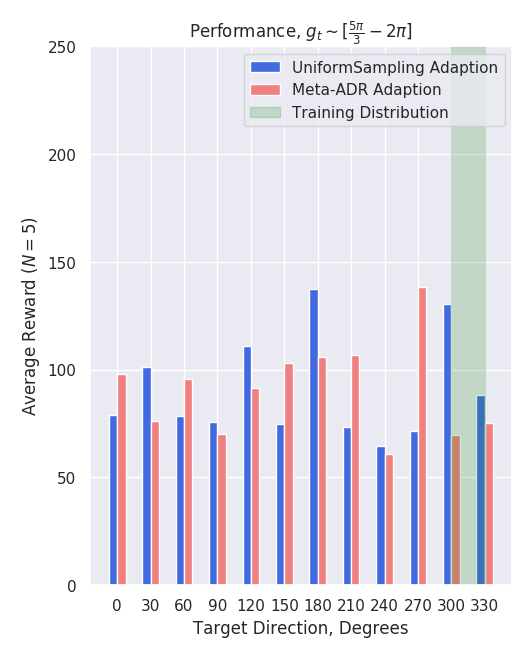}
    \caption{In complex, high-dimensional environments, training task distributions can wildly vary performance. Even in the Humanoid Directional task, Meta-ADR allows MAML to generalize across the range, although it too is affected in terms of total return when compared to the same algorithm trained with "good" task distributions.}
    \label{fig:humanoid-results}
\end{figure}

When studying the generalization capabilities on this difficult continuous control task, we are particularly interested in \textit{symmetric} versions of the task; for example, tasks that sample the right and left semi-circles of the task space. We repeat this experiment with many variants of this symmetric task description, and report representative results due to space in Figure \ref{fig:humanoid-results}.

When testing various training distributions, we find that, in general, learned curricula stabilize the algorithm. We see more consistent performance increases, with smaller losses in performance in the directions that \textit{UniformSampling-MAML} outperforms the learned curriculum. However, as noted in Tables \ref{table:humanoid-evals1} and \ref{table:humanoid-evals2}, we see that again, the task distribution $p(\tau)$ is an extremely sensitive hyperparameter, causing large shifts in performance when uniformly sampling from those ranges. Worse, this hyperparameter seems to cause \textit{counter-intuitive} gains and drops in performance, both on in \textit{and} out-of-distribution tasks.

\begin{table}[]
\small
\caption{We compare agents trained with random curricula on different but symmetric task distributions $p(\tau)$. Changing the distribution leads to counter-intuitive drops in performance on tasks both in- and out-of-distribution. }
\begin{tabular}{cccc}
p($\tau$)                                        & $\theta=0$              & $\theta=30$            & $\theta=60$             \\ \hline
\multicolumn{1}{c|}{\textit{0 - 360}} & \textit{62.39$\pm$7.01} & \textit{64.7$\pm$5.38} & \textit{99.93$\pm$77.5} \\ \cline{2-4} 
\multicolumn{1}{c|}{0 - 60}                      & 71.51$\pm$16.74         & 82.95$\pm$32.06        & 95.96$\pm$37.49         \\
\multicolumn{1}{c|}{0 - 180}                     & 171.77$\pm$117.8        & 221.4$\pm$91.66        & 87.78$\pm$45.41         \\ \cline{2-4} 
\multicolumn{1}{c|}{300 - 360}                   & 65.64$\pm$10.42         & 95.21$\pm$40.08        & 105.4$\pm$50.75         \\
\multicolumn{1}{c|}{180 - 360}                   & 134.52$\pm$70.07        & 79.69$\pm$26.01        & 59.52$\pm$2.73         
\end{tabular}

\label{table:humanoid-evals1}
\end{table}

\begin{table}[]
\small
\caption{Evaluating tasks that are \textit{qualitatively} similar, for example running at a heading offset from the starting heading by 30 degrees to the \textit{left or right}, leads to different performances from the same algorithm.}
\begin{tabular}{cccc}
p($\tau$)                                        & $\theta=180$               & $\theta=330$             & $\theta=300$             \\ \hline
\multicolumn{1}{c|}{\textit{0 - 360}} & \textit{154.1$\pm$121.6} & \textit{81.26$\pm$16.39} & \textit{75.47$\pm$18.92} \\ \cline{2-4} 
\multicolumn{1}{c|}{0 - 60}                      & 120.2$\pm$62.74           & 84.34$\pm$35.9           & 126.2$\pm$101.3        \\
\multicolumn{1}{c|}{0 - 180}                     & 87.78$\pm$45.41            & 103.5$\pm$87.24          & 89.49$\pm$31.0           \\ \cline{2-4} 
\multicolumn{1}{c|}{300 - 360}                   & 116.03$\pm$52.44           & 81.25$\pm$43.82          & 100.45$\pm$48.41         \\
\multicolumn{1}{c|}{180 - 360}                   & 99.1$\pm$88.85             & 80.52$\pm$21.79          & 80.67$\pm$21.82         
\end{tabular}

\label{table:humanoid-evals2}

\end{table}

While learned curricula seem to help in such a task, a more important consideration from many of these experiments is the variance in performance \textit{between} tasks. 
As \textit{generalization} across evaluation tasks is a difficult metric to characterize due to the inherent issues when \textit{comparing} methods, it is tempting to take the best performing tasks, or average across the whole range.  However, as we show in the remaining sections, closer inspection on each of the above experiments sheds light on major issues with the evaluation approaches standard in the meta-RL community today.

\subsection{Failure Cases of MAML}
\label{sec:failurecases}
In this section, we discuss intermediate and auxillary results from each of our previous experiments, highlighting uninterpretable algorithm bias, meta-overfitting, and performance benchmarking in meta-RL. 

\subsubsection{Non-Uniform Generalization}
\label{sec:nonuniform-gen}

\begin{figure}[tb]
    \centering
    \includegraphics[width=1.0\columnwidth]{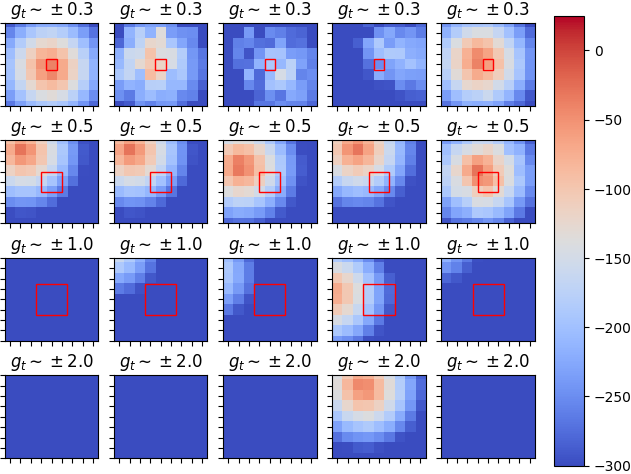}
    \caption{Uniform sampling causes MAML to show bias towards certain tasks, with the effect being compounded with instability when using "bad" task distributions, here shown as $\pm 0.3, \pm 1.0, \pm 2.0$ in the 2D-Navigation-Dense environment.}
    \label{fig:bias-2d}
\end{figure}

To readers surprised by the poor generalization capabilities of MAML on such a simple task seen in Figure \ref{fig:motivation2d}, we offer Figure \ref{fig:bias-2d}, an unfiltered look at each seed used to calculate each image in Figure \ref{fig:motivation2d}. 

What we immediately notice is the high variance in all but the standard variant of the task, an agent trained on goals with coordinates sampled from $g_t \sim  [-0.5, 0.5]$. We even see a reoccurring bias towards certain tasks (visualized as the \textit{top-left} of the grid). Interestingly, when changing the uniform sampling to a learned curriculum, we no longer see such high-variance in convergence across tasks. While our results seem in opposition to many works in the meta-reinforcement learning area, we restate that in our setting, we can only evaluate the \textit{final} policy, as the notion of \textit{best}-performing loses much of its meaning when evaluating for generalization.

\subsubsection{Meta-Overfitting}
\label{sec:meta-overfitting}

Many works in the meta-reinforcement learning setting focus on final adaption \textit{performance}, but few works focus on the \textit{loss of performance} after the adaption step. Coined by \citep{deleunegativeadapt19} as \textit{negative adaption}, the definition is simple: the loss in performance \textit{after} a gradient step at meta-test time. Negative adaptation occurs when a pre-adaption policy has overfit to a particular task during meta-training. During meta-test time, an additional gradient steps degrade performance, leading to negative adaptation. 

We extensively evaluate negative adaption in the \textit{Humanoid-Direc-2D} benchmark described in Section \ref{sec:humanoiddirectional-results}, providing correlation results between performance and the \textit{difference} between the post- and pre-adaption performance. 

When we systematically evaluate negative adaption across all tested \textit{Humanoid-Direc-2D} training distributions, we notice an interesting correlation between performance and the amount of negative-adaptation. Both methods produce near-linear relationships between the two quantities, but when evaluating generalization, we need to focus on the left-hand side of the x-axis, where policies already are performing poorly, and what qualitative effects extra gradient steps have.

\begin{figure}[tb]
    \centering
    \includegraphics[width=1\columnwidth]{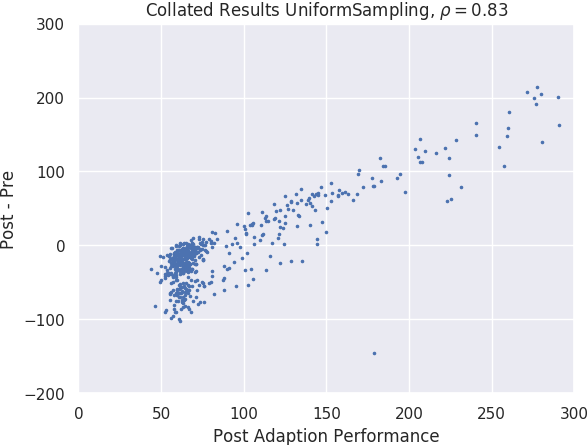}
    \caption{When we correlate the final performance with the amount of quality of \textit{adaption}, we see a troubling trend. MAML seems to overfit to certain tasks, with many tasks that were already neglected during training showing worse post-adaptation returns.}
    \label{fig:humanoid-correlation}
\end{figure}

We notice a characteristic sign of \textit{meta-overfitting}, where strongly performing policies continue to perform well, but poorly performing ones stagnate, or more often, degrade in performance. When tested, Meta-ADR does not help in this regard, despite having slightly stronger final performance in tasks.

\subsubsection{Random Sampling and Performance Improvements}
\label{sec:randomsampling}

\begin{figure}[ht]
    \centering
    \includegraphics[width=1\columnwidth]{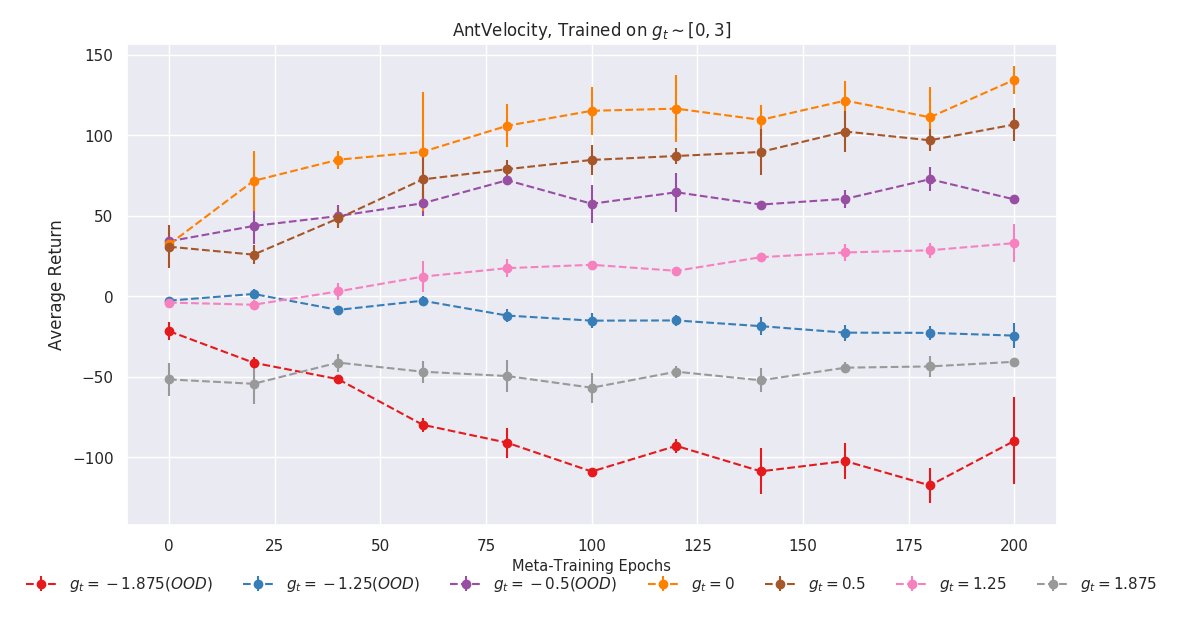}
    \caption{When individually plotting out each target velocity, we see a strong bias towards the easier variants of the task. Only the easier variants of the task produce monotonically increasing learning curves.}
    \label{fig:antvel-convergence}
\end{figure}

Most alarmingly, we present results on the \textit{evaluation} of tasks, particularly in the Locomotion Velocity tasks. As noted in \citep{deleunegativeadapt19}, we see a characteristic \textit{pyramid} when evaluating generalization of locomotion agents trained to achieve a target task. However, in many papers concerning meta-RL, we see monotonic growth curves in performance on such environments. In Figure \ref{fig:antvel-convergence}, we show the issues in reporting such curves.

When we sample uniformly from the same distribution that we train on, say target velocities pulled from $v_t \sim [0, 3]$, the evaluation sampling can highly affect our results. We evaluate the training curves of the MAML agent, and see that only those closest to the easiest variant of the task ($v_t = 0$, where an agent must stand still) produce monotonically increasing learning curves. More interestingly,  target velocities closer to zero but \textit{out-of-distribution} (OOD) show better performance than larger target velocities that are in distribution. As the deviation away from a target velocity of 0 becomes larger, the learning curves stagnate or even start to degrade.

Unlike the previous two sections, which stem interesting research directions such as \textit{why does MAML show bias towards certain tasks} and \textit{how can we fix negative adaption}, our analysis here points towards a fundamental flaw in the \textit{task} design of the Target Velocity Locomotion tasks commonly used in Meta-RL benchmarking. 

\section{Conclusion}

We present \textit{Meta-ADR}, a curriculum learning algorithm suitable for helping gradient-based meta-learners generalize better in a meta-reinforcement learning setting. We show a strong dependence between the performance of MAML and the correct task distribution. When switching out only the random sampling of tasks for such a learned curriculum, we show strong performance across a variety of meta-RL benchmarks. From our experiments, we highlight issues with current meta-RL benchmarking, focusing on a need for \textit{generalization} evalution, a proper, exclusive train-test task separation, and better evaluation tasks in general.

\subsection*{Acknowledgements}

The authors gratefully acknowledge the Natural Sciences and Engineering Research Council of Canada (NSERC), the Fonds de Recherche Nature et Technologies Quebec (FQRNT), Calcul Quebec, Compute Canada, the Canada Research Chairs, Canadian Institute for Advanced Research (CIFAR) and Nvidia for donating a DGX-1 for computation. BM would like to thank Glen Berseth for helpful discussions in early drafts of this work, and IVADO for financial support.